\documentclass[letterpaper, 10 pt, conference]{ieeeconf}  

\IEEEoverridecommandlockouts                              

\usepackage{graphicx}
\usepackage{amsmath,amsthm,amssymb}        
\usepackage[nolist]{acronym}
\usepackage{calrsfs}
\usepackage[capitalise]{cleveref}
\usepackage{placeins}
\usepackage{subcaption}
\usepackage{algorithm}
\usepackage{algorithmicx}
\usepackage{mathtools}      
\usepackage{diagbox}
\usepackage{multirow}
\usepackage{verbatim}
\usepackage{color}
\usepackage{xcolor}
\usepackage[normalem]{ulem}
\usepackage{balance}
\if@cref@capitalise
\crefname{theorem}{TheoUpperCase}{TheoUppercasesS}
\else
\crefname{theorem}{theoLowercase}{theoLowercaseS}
\fi

\newcommand{\todo}[1]{{\color{red}{{#1}}}} 
\newcommand{\BF}[1]{{\color{blue}{{#1}}}} 
\newcommand{\LF}[1]{{\color{magenta}{{#1}}}} 
\DeclareMathAlphabet{\pazocal}{OMS}{zplm}{m}{n}
\newcommand{\A}{\pazocal{A}}
\newcommand{\B}{\pazocal{B}}
\newcommand{\order}{\pazocal{O}}
\newcommand{\U}{\pazocal{U}}
\newcommand{\W}{\pazocal{W}}

\theoremstyle{plain}
\newtheorem{lem}{\textbf{Lemma}}

\usepackage{algorithm}
\usepackage{algpseudocode}

\title{\LARGE \bf
Model Predictive Contouring Control for Collision Avoidance in Unstructured Dynamic Environments
}
 
\author{Bruno Brito, Boaz Floor, Laura Ferranti and Javier Alonso-Mora
\thanks{This work was supported by the Amsterdam Institute for Advanced Metropolitan Solutions and the Netherlands Organisation for Scientific Research (NWO) domain Applied Sciences.}
\thanks{B. Brito and B. Floor contributed equally}
\thanks{The authors are with the Cognitive Robotics (CoR) department,
        Delft University of Technology, 2628 CD Delft, The Netherlands
    {\tt\small \{bruno.debrito,l.ferranti, j.alonsomora\}@tudelft.nl}}%
}

\begin{document}
   
\begin{acronym}
	\acro{agv}[AGV]{Autonomous Ground Vehicle}%
	\acro{gvo}[GVO]{Generalized Velocity Obstacle}%
	\acro{lmpcc}[LMPCC]{Local Model Predictive Contouring Control}%
	\acro{nmpcc}[NMPCC]{Nonlinear MPCC}%
	\acro{kkt}[KKT]{Karush-Kuhn-Tucker}
	\acro{mpc}[MPC]{Model Predictive Control}%
	\acro{mpcc}[MPCC]{Model Predictive Contouring Control}%
	\acro{rrt}[RRT]{Rapidly-exploring Random Tree}%
 	\acro{vo}[VO]{Velocity Obstacle}%
\end{acronym}


\maketitle
\thispagestyle{empty}
\pagestyle{empty}

\begin{abstract}
This paper presents a method for local motion planning in unstructured environments with static and moving obstacles, such as humans. Given a reference path and speed, our optimization-based receding-horizon approach computes a local trajectory that minimizes the tracking error while avoiding obstacles. We build on nonlinear model-predictive
contouring control (MPCC) and extend it to incorporate a static map by computing, online, a set of convex regions in free space. We model moving obstacles as ellipsoids and provide a correct bound to approximate the collision region, given by the Minkowsky sum of an ellipse and a circle. 
Our framework is agnostic to the robot model. We present experimental results with a mobile robot navigating in indoor environments populated with humans. Our method is executed fully onboard without the need of external support and can be applied to other robot morphologies such as autonomous cars.
\end{abstract}

\section{INTRODUCTION}

Applications where autonomous ground vehicles (AGVs) closely navigate with humans in complex environments require the AGV to safely avoid static and moving obstacles while making progress towards its goal.
Motion planning and control for AGVs are typically addressed as two independent problems \cite{Schwarting2018}. In particular, the motion planner generates a collision-free path and the motion controller tracks such a path by directly commanding the AGV's actuators.
Our method combines both motion planning and control in one module, relying on constrained optimisation techniques, to generate kinematically feasible local trajectories with fast replanning cycles.
More specifically, we rely on a Model Predictive Controller (MPC) to compute an optimal control command for the controlled system, which directly incorporates the predicted intentions of dynamic obstacles. Consequently, it reacts in advance to smoothly avoid moving obstacles.
We propose a practical reformulation of the \ac{mpcc} approach~\cite{schwarting2017safe}, namely, a \ac{lmpcc} approach, suitable for real-time collision-free navigation of AGVs in complex environments with several agents. Our design is fully implemented on board of the robot including localization and environment perception, i.e., detection of static obstacles and pedestrians. Our design runs in real time thanks to its lightweight implementation. The method relies on an open-source solver \cite{Houska2011a} and will be released with this paper as a ROS module. Our method can be adapted to other robot morphologies, such as cars.
\begin{figure}
\centering
\includegraphics[width=.85\columnwidth]{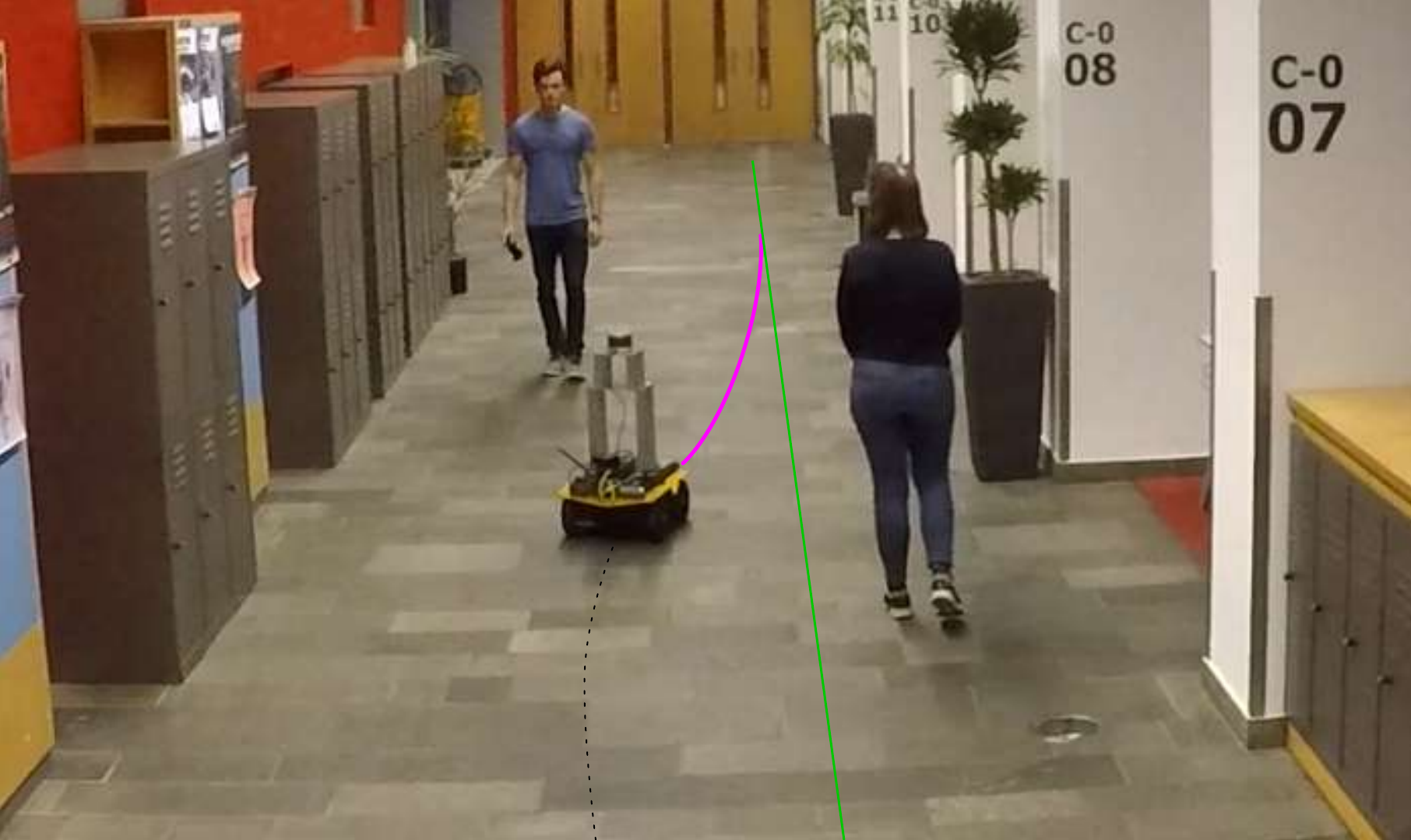}
\caption{The faculty corridor was the scenario used to evaluate the capabilities of our method to avoid dynamic obstacles.}
\label{fig:front_page}
\end{figure}

\subsection{Related Work}

Collision avoidance in static and dynamic environments can be achieved via reactive methods, such as time-varying artificial potential fields~\cite{Khatib1986}, the dynamic window~\cite{fox1997dynamic}, social forces~\cite{Ferrer2013} and velocity obstacles \cite{AlonsoMora:2018ur}.
Although these approaches work well in low speed scenarios, or scenarios of low complexity, they produce highly reactive behaviours.
More complex and predictive behaviour can be achieved by employing a motion planner.
 Our method relies on model predictive control (MPC)~\cite{maciejowski2002predictive,borrelli2017predictive} to obtain smooth collision-free trajectories that optimize a desired performance index, incorporate the physical constraints of the robot and the predicted behavior of the obstacles.

Due to the complexity of the motion planning problem, path planning and path following were usually considered as two separate problems. 
Many applications of MPC for path-following control are found in the literature, e.g.,~\cite{Borrelli2005, Chang2017}. These methods assume the availability of a collision-free smooth path to follow. In contrast, we use MPC for local motion planning and control in the presence of static and dynamic obstacles.

Several approaches exist for integrated path following and control for dynamic environments. These include: 
employing a set of motion primitives and optimizing the control commands to execute them \cite{Howard2010} and offline computation of tracking error bounds via reachability analysis that ensures a safety region around the robot for online planning~\cite{fridovich2017planning,Fisac2018}. These approaches, however, do not allow to incorporate the predicted intentions of the dynamic obstacles and consequently, can lead to reactive behaviors. 
To overcome the latter issue, \cite{Lam2010} proposed the model predictive contouring control (MPCC) that allows to explicitly penalize the deviation from the path (in terms of contouring and lateral errors) and include additional constraints. Later, \cite{Brown2017} designed a MPC for path following within a stable handling envelope and an environmental envelope. While \cite{Brown2017} is tailored to automotive applications without dynamic obstacles, MPCC has been employed to handle static and dynamic obstacles in structured driving scenarios \cite{schwarting2017safe}. 
There, static collision constraints were formulated as limits on the reference path and thus limited to on-the-road driving scenarios.
The previous approaches do not account for the interaction effects between the agents and may fail in crowded scenarios, a problem known as the freezing robot problem (FRP) \cite{kruse2013human}. Interaction Gaussian Processes (IGP) \cite{trautman2017sparse} can be used to model each individual's path. The interactions are modeled with a nonlinear potential function. The resulting distribution, however, is intractable, and sampling processes are required to approximate a solution, which requires high computational power and is only real-time for a limited number of agents.
Learning-based approaches address this issue by learning the collision-avoidance strategy directly from offline simulation data \cite{saferl}, or the complex interaction model from raw sensor readings \cite{long2018towards}. Yet, both methods learn a reactive collision avoidance policy and do not account for the kinodynamic constraints of the robot.

Without explicitly modeling interaction, we propose a method for local motion planning, which is real-time, incorporates the robot constraints and optimizes over a prediction horizon. Our approach relies on MPCC and extends it to mobile robots operating in unstructured environments. Similar to \cite{AlonsoMora:2017kp}, among others, we employ polyhedral approximations of the free space, which can provide larger convex regions than safety bubbles~\cite{Zhu2015}.

\subsection{Contribution}
We build on \cite{schwarting2017safe}, with the following contributions to make the design applicable to mobile robots navigating in unstructured environments with humans:
\begin{itemize}
\item A static obstacle avoidance strategy that explicitly constrains the robot's positions along the prediction horizon to a polyhedral approximation of the collision-free area around the robot.
\item A closed-form bound to conservatively approximate collision avoidance constraints that arise from ellipsoidal moving obstacles.
\item A fully integrated MPCC approach that runs in real-time on-board of the robot and with on-board perception.
\end{itemize}
We present experimental results with a mobile robot navigating in indoor environments among static and moving obstacles and compare them with three state-of-art planners, namely the dynamic window \cite{fox1997dynamic}, a classical MPC for tracking~\cite{Chang2017} and a socially-aware motion planner \cite{everett2018motion}. Finally, to illustrate the generality of the proposed MPCC framework, we present results with a simulated car.


\section{PRELIMINARIES}\label{sec:problem_formulation}
\subsection{Robot description}\label{sec:robot}
Let $B$ denote an AGV on the plane $\W = \mathbb{R}^2$. The AGV dynamics are described by the discrete-time nonlinear system
\begin{equation}\label{eq:state_dynamics}
\boldsymbol{z}(t+1) = f(\boldsymbol{z}(t),\boldsymbol{u}(t)),
\end{equation}
where $\boldsymbol{z}(t)$ and $\boldsymbol{u}(t)$ are the state and the input of the robot, respectively, at time $t\geq 0$\footnote{In the remainder of the paper we omit the time dependency when it is clear from the context.}. For the case of our mobile robot we consider the state to be equal to the configuration, that is, $\boldsymbol{z}(t)\in \pazocal{C} = \mathbb{R}^2 \times \pazocal{S}$.
For the case of a car (Sec.~\ref{sec:results_simulations}), the state space includes the speed of the car. 
The area occupied by the robot at state $\boldsymbol{z}$ is denoted by $\pazocal{B}(\boldsymbol{z})$. which is approximated by a union of $n_c$ circles, i.e., $\pazocal{B}(\boldsymbol{z}) \subseteq \bigcup_{c \in \{1,\dots,n_c \}} \pazocal{B}_c(\boldsymbol{z}) \subset \W$.
The center of each circle, in the inertial frame, is given by 
$\boldsymbol{p} + {R}_B^W(\boldsymbol{z}) \boldsymbol{p}_c^B$. Where $\boldsymbol{p}$ is the position of the robot (extracted from $\boldsymbol{z}$), ${R}_B^W(\boldsymbol{z})$ is the rotation matrix given by the orientation of the robot, and $\boldsymbol{p}_c^B$ is the center of circle $c$ expressed in the body frame.

\subsection{Static obstacles}
The static obstacle environment is captured in an occupancy grid map, where
the area occupied by the static obstacles is denoted by $\pazocal{O}^{\textrm{static}} \subset \pazocal{W}$.
In our experiments we consider both a global map, which is built a priori and used primarily for localization, and a local map from the current sensor readings. Therefore, the static map is continuously updated locally. Dynamic obstacles, such as people, that are recognized and tracked by the robot are removed from the static map and considered as moving obstacles.
\subsection{Dynamic obstacles}
Each moving obstacle $i$ is represented by an ellipse of area $\A_i \subset \pazocal{W}$, defined by its semi-major axis $a_i$, its semi-minor axes $b_i$, and a rotation matrix $R_i(\psi)$. We consider a set of moving obstacles $i \in \pazocal{I} := \{1, \dots , n\}$, where $n$ can vary over time. The area occupied by all moving obstacles at time instant $t$ is given by $\pazocal{O}_t^{\textrm{dyn}} = \bigcup_{i \in \{1,\dots,n \}} \A_i(\boldsymbol{z}_i(t))$, where $\boldsymbol{z}_i(t)$ denotes the state of moving obstacle $i$ at time $t$. In this work, and without a loss of generality, we assume a constant velocity model with Gaussian noise $\mathbf{\omega}_o(t) \sim \mathcal{N}(0, Q_o(t))$ in acceleration, that is, $\ddot {\mathbf{p}}_i(t) = \mathbf{\omega}_i(t)$, where $\mathbf{p}_i(t)$ is the position of obstacle $i$ at time $t$. Given the measured position data of each obstacle, we estimate their future positions and uncertainties with a linear Kalman filter~\cite{Welch1995}.


\subsection{Global Reference Path}
\label{sec:preliminaries_global_reference_path}
Consider that a reference path is available. In its simplest form, the reference path can be a straight line to the goal position or a straight line in the direction of preferred motion. But it could also be given by a global planner. We consider a global reference path $\pazocal{P}$ consisting of a sequence of path segments connecting $M$ way-points $\boldsymbol{p}^{r}_m = [x^p_m,y^p_m] \in \W$ with $m \in \pazocal{M} := \{1,\dots,M\}$. For smoothness, we consider that each path segment $\varsigma_m(\theta)$ is defined by a cubic polynomial. We denote by $\theta$ a variable that (approximately) represents the traveled distance along the reference path, and which is described in more detail in Sec. \ref{sec:reference_path}. We do not require the reference to be collision free, therefore, the robot may have to deviate from it to avoid collisions.

\subsection{Problem Formulation}
The objective is to generate, for the robot, a collision free motion for $N$ time-steps in the future, while minimizing a cost function $J$ that includes a penalty for deviations from the reference path. This is formulated in the optimization problem
\begin{subequations}
\label{eq-problem}
\begin{align}
J^* = & \min_{\boldsymbol{z}_{0:N},\boldsymbol{u}_{0:N-1},\theta_{0:N-1}} \sum^{N-1}_{k =0} J(\boldsymbol{z}_k, \boldsymbol{u}_k, \theta_k) + J(\boldsymbol{z}_N,\theta_N)\nonumber \\
\label{eq:dynamics_path_parameter}\text{s.t.} \quad & \boldsymbol{z}_{k+1} = f(\boldsymbol{z}_{k},\boldsymbol{u}_{k}), \,\,
\theta_{k+1} = \theta_{k} + v_{k} \tau,  \\
& \B(\boldsymbol{z}_k) \cap \big( \pazocal{O}^{\textrm{static}} \cup \pazocal{O}_k^{\textrm{dyn}} \big) = \emptyset,\\
&\label{eq:state_control_constraints} \boldsymbol{u}_k \in \U, \,\, \boldsymbol{z}_{k} \in \pazocal{Z},\, \, \boldsymbol{z}_{0}, \theta_0 \textnormal{ given}.
\end{align}
\end{subequations}
Where $v_k$ is the forward velocity of the robot (for the mobile robot it is part of the input and for the car it is part of the state), $\tau$ is the time-step and
$\pazocal{Z}$ and $\U$ are the set of admissible states and inputs, respectively. $\boldsymbol{z}_{1:N}$ and $\boldsymbol{u}_{0:N-1}$ are the set of states and control inputs, respectively, over the prediction horizon $T_{\textrm{horizon}}$, which is divided into $N$ prediction steps. $\theta_k$ denotes the predicted progress along the reference path at time-step $k$. By solving the optimization problem, we obtain a locally optimal sequence of commands $[\boldsymbol u_t^*]_{t=0}^{t=N - 1}$ to guide the robot along the reference path while avoiding collisions with static and moving obstacles. 


\section{METHOD}\label{sec:method}
The proposed method consists of the following steps, which are executed in every planning loop.

\paragraph*{1} Search for a collision-free region in the updated static map centered on the robot and constrain the control problem such that the robot remains inside (\cref{sec:static_collision_avoidance}).
\paragraph*{2} Predict the future positions of the dynamic obstacles and use the corrected bound to ensure dynamic collision avoidance (\cref{sec:dynamic_collision_avoidance}).
\paragraph*{3} Solve a modified MPCC formulation applicable to mobile robots (\cref{sec:lmpcc}).

\subsection{Static collision avoidance}\label{sec:static_collision_avoidance}



Given the static map of the environment, we compute a set of convex four-sided polygons in free space. This representation can provide larger collision-free areas compared with other approximations such as circles. 

To obtain the set of convex regions at time $t$, we first shift the optimal trajectory computed at time $t-1$, namely, $\boldsymbol{q}_{0:N}= [\boldsymbol{p}^*_{1:N|t-1}, \boldsymbol{q}_N]$, where $\boldsymbol{q}_N$ is a extrapolation of the last two points, that is, $\boldsymbol{q}_N = 2 \boldsymbol{p}^*_{N|t-1}-\boldsymbol{p}^*_{N-1|t-1}$.
\begin{figure}[t!]
\centering
\includegraphics[trim={2.2cm 0cm 0cm 7.5cm},clip ,width=.8\columnwidth]{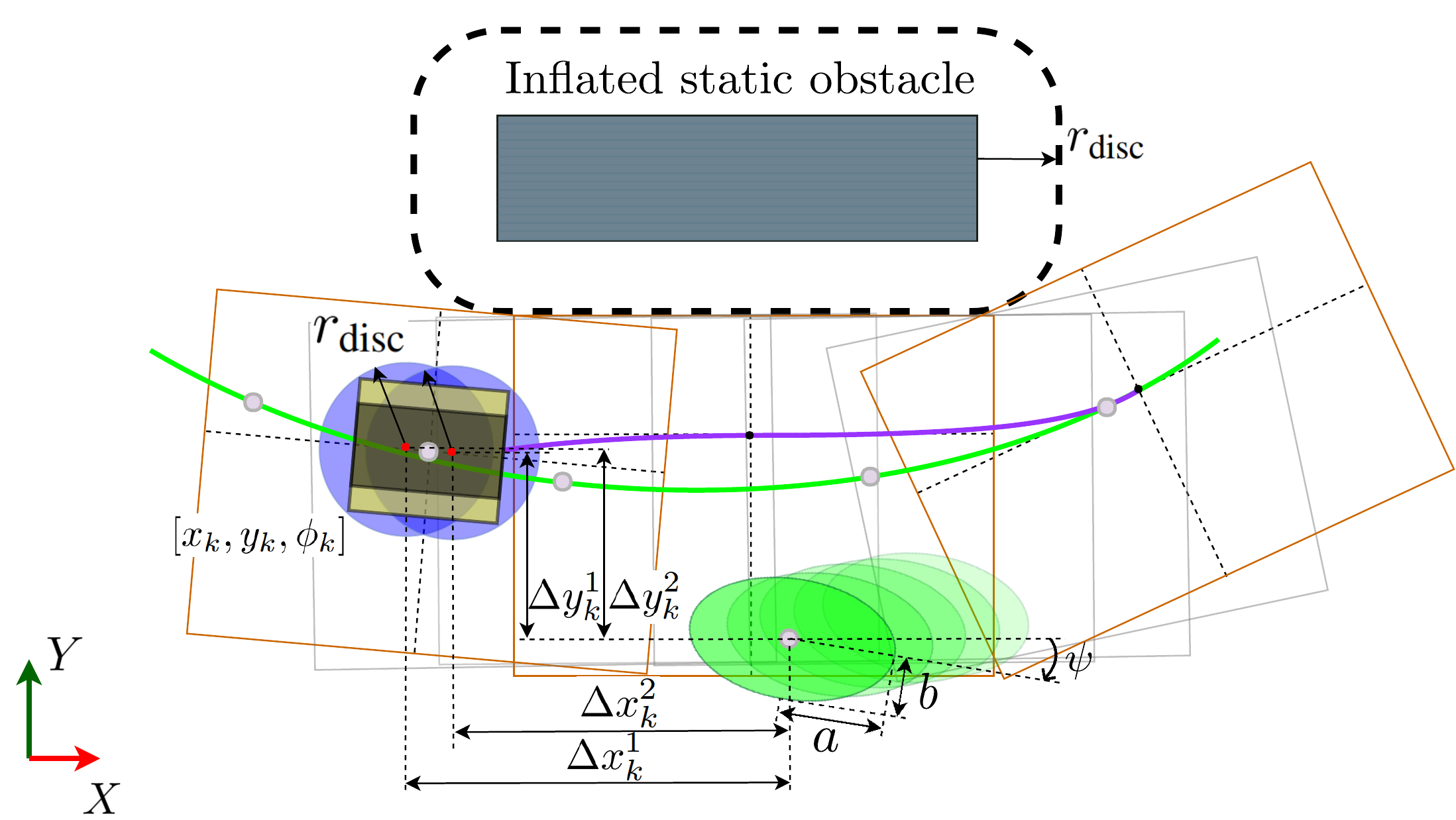}
\caption{Representation of the convex free space (orange squares) around each prediction step on the prediction horizon (purple line) with respect to the inflated static environment and the collision space of the dynamic obstacle (green ellipses) with respect to the vehicle representing discs (blue).}
\label{fig:obstacles}
\end{figure}
Then, for each point $\boldsymbol{q}_k$ ($k=1,..., N$) we compute a convex region in free space, given by a set of four linear constraints $c^{\text{stat}}_k(\boldsymbol{p}_k) = \bigcup^{4}_{{l} = 1} c^{\text{stat},l}_k(\boldsymbol{p}_k)$. This region separates $\boldsymbol{q}_k$ from the closest obstacles. In our implementation we compute a rectangular region aligned with the orientation of the trajectory at $\boldsymbol{q}_k$, where each linear constraint is obtained by a search routine and reduced by the radius of the robot circles $r_{\textrm{disc}}$.  
Figure~\ref{fig:obstacles} shows the collision-free regions along the prediction horizon defined as yellow boxes.
At prediction step $k$ for a robot with state $\boldsymbol{z}$ the resulting constraint for disc $j$ and polygon side $l$ is
 \begin{align}\label{eq:env_constraint2}
\begin{split}
    c^{\text{stat},l,j}(\boldsymbol{z}) = h^l \, -& \, \vec{n}^l \cdot \big(\boldsymbol{p} - \,\, {R}_B^W(\boldsymbol{z})\boldsymbol{p}_{j}^B \big) 
    > 0,
\end{split}	
\end{align}
where {$h^l$} and $\vec{n}^l$ define each side of the polygons.

\subsection{Dynamic collision avoidance}\label{sec:dynamic_collision_avoidance} 
Recall that each moving obstacle $i$ is represented by its position $\boldsymbol{p}_i(t)$ and an ellipse of semi-axis $a_i$ and $b_i$ and a rotation matrix $R_i(\psi)$. For each obstacle $i \in \{1,\dots,n\}$, and prediction step $k$, we impose that each circle $j$ of the robot does not intersect with the elliptical area occupied by the obstacle. Omitting $i$ for simplicity, the inequality constraint on each disc of the robot with respect to the obstacles is
\begin{equation}\label{eq:ineq_constraint}
\begin{split}
\!\!\!c_k^{\textrm{obst},j}(\boldsymbol{z}_k) \!\!= \!\!\begin{bmatrix}
\Delta x_k^j\\
\Delta y_k^j
\end{bmatrix}^\textrm{T}\!\!\!\!
R(\psi)^\textrm{T}\!\!\begin{bmatrix}
\frac{1}{\alpha^2} & 0\\ 0 & \frac{1}{\beta^2}
\end{bmatrix}\!\! R(\psi)\!\!\!\ \begin{bmatrix}
\Delta x_k^j\\
\Delta y_k^j
\end{bmatrix}
> 1,
\end{split}
\end{equation}
where the distance between disc $j$ and the obstacle is separated into its $\Delta x^j$ and $\Delta y^j$ components (\cref{fig:obstacles}). The parameters $\alpha$ and $\beta$ are the semi-axes of an enlarged ellipse that includes the union of the original ellipse and the circle. 

\begin{figure}[t!]
\centering 
\includegraphics[width=0.6\columnwidth]{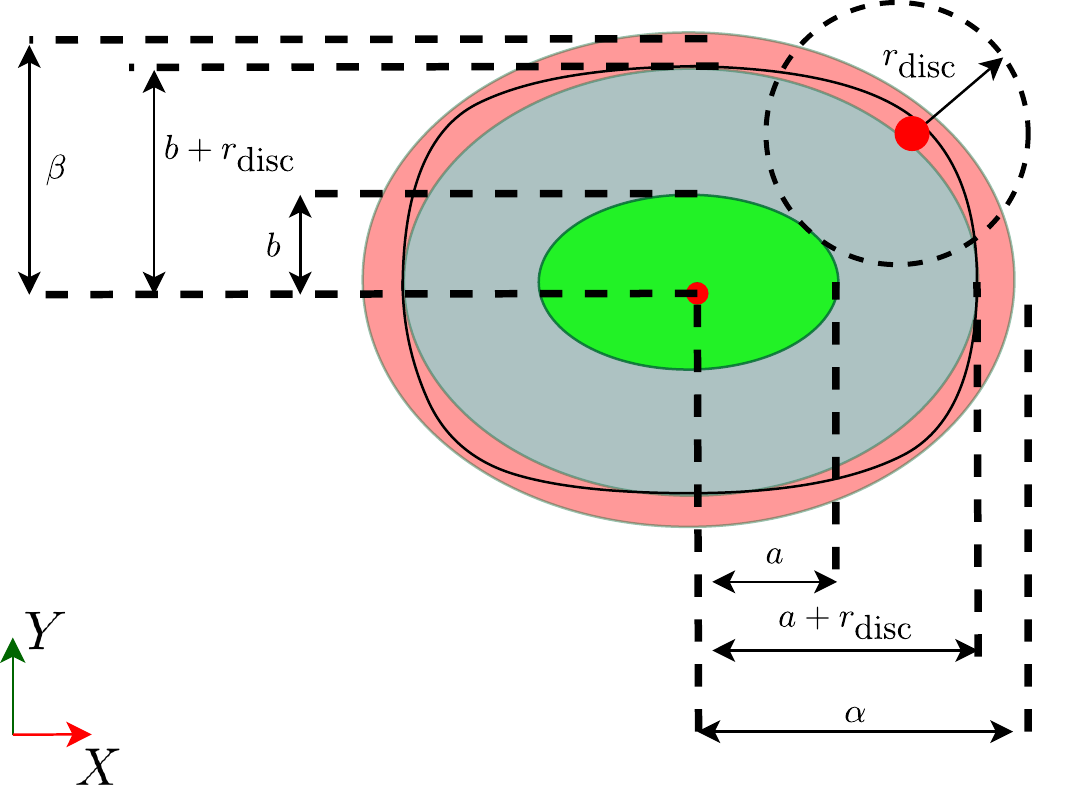}
\caption{Safety boundary computation. The ellipsoid with semi-major axis and semi-minor axis, $a$ and $b$ respectively, is represented in green, the ellipsoid with axis enlarged by $r_{\textrm{disc}}$ is represented in gray, the Minkowsky sum is represented in black and the ellipsoid enlarged by $\delta$ is represented in red.}
\label{fig:bound}
\end{figure}

While previous approaches approximated the Minkowski sum of the ellipse with the circle as an ellipse of semi-major $\alpha = a + r_{\text{disc}}$ and semi-minor axis $\beta = b + r_{\text{disc}}$ \cite{schwarting2017safe}, this assumption is not correct and collisions can still occur \cite{chen2017constrained}.
We now describe how to compute the values for $\alpha$ and $\beta$ such that collision avoidance is guaranteed, represented by the larger in light red ellipsoid in Fig. \ref{fig:bound}. 

Consider two ellipsoids $E_1= \operatorname{Diag}(\frac{1}{a^2}\frac{1}{b^2})$ and $E_2 = \operatorname{Diag}\left(\frac{1}{(a+\delta)^2}, \frac{1}{(b+\delta)^2}\right)$. $E_1$ is an ellipsoid with $a$ and $b$ as semi-major and semi-minor axes, respectively. $E_2$ represents the ellipsoid $E_1$  enlarged by $\delta$ in both axis. The goal is to find the smallest ellipsoid that bounds the Minkowsky sum. This is equivalent to find the minimum value of $\delta$ such that the minimum distance between ellipsoid $E_1$ and $E_2$ is bigger than {$r$}\footnote{Note we use $r$ instead of $r_{\textrm{disc}}$ in the reminder of the section to simplify the notation.}, the radius of the circle bounding the robot.

\begin{lem}[\cite{uteshev2015metric}]\label{eq:minimum_polynom}
        Let $X^\textrm{T}A_1X=1$ and $X^\textrm{T}A_2X=1$ be two quadratics in $\mathrm{R}^n$. Iff the matrix $A_1-A_{2}$ is sign definite, then the square of the distance between the quadratic $X^\textrm{T}A_1X=1$ and the quadratic $X^\textrm{T}A_2X=1$ equals the minimal positive zero of the polynomial.
        \begin{align*}
                 F(z) = D_\lambda(\det{\lambda A_1+(z-\lambda A_2)-\lambda(z-\lambda A_1A_2)})
        \end{align*}
        where $D$ stands for the discriminant of the polynomial treated with respect to $\lambda$.
\end{lem}

Considering $A_1=E_1$, $A_2=E_2$ and $\{\delta, a,b,\} \in \mathrm{R}^+$ this ensures that $E_1-E_{2}$ is sign definite. Hence, we can apply Lemma \ref{eq:minimum_polynom} to determine the polynomial $F(z)$ and its roots. For the two ellipsoids $E_1$ and $E_2$, the roots {$\lambda$} of $F(z)$ are:

{\begin{equation}
    \lambda \in \left\{
\begin{aligned}
 \frac{(2a(r+\delta)^3 + 2b(r+\delta)^3 + 4ab(r+\delta)^2}{(a^2 + 2ab + 2r a + b^2 + 2r b}, \\ 
(r+\delta)^2, \\ 
4a^2 + 4a(r+\delta) + (r+\delta)^2, \\ 
4b^2 + 4b(r+\delta) + (r+\delta)^2
\end{aligned}
\right\}.
\end{equation}}
The first two roots have multiplicity two. The minimum distance equation is the square root of the minimal positive zero of $F(z)$. Thus, the minimum enlargement factor is found by solving for the value of $\delta$ that satisfies
$ \min_j \lambda(j) = r^2 $.

A closed form formula can be obtained by noting that the $\min_j \lambda(j)$ is achieved for the first root and solving this equation. Due to its length, it is not presented in this paper but can be found at \footnote{A Mathemathica notebook with the derivation of the bound and a Matlab script as example of its computation can be found in
http://www.alonsomora.com/docs/19-debrito-boundcomputation.zip.}.
This value of semi-axis $\alpha = a+\delta$ and $\beta = b+\delta$ guarantee that the constraint ellipsoid entirely bounds the collision space.



\subsection{Model Predictive Contouring Control}\label{sec:lmpcc} 
MPCC is a formulation specially tailored to path-following problems. This section presents how to modify the baseline method \cite{schwarting2017safe} and make it applicable to mobile robots navigating in unstructured environments with on-board perception. 

\subsubsection{Progress on reference path}\label{sec:path_parameter}

Eq. \ref{eq-problem} approximates the evolution of the path parameter by the travelled distance of the robot. In each planning stage we initialize \BF{$\theta_0$}. We find the closest path segment, denoted by $m$, and compute the value of $\theta_0$ via a line search in the neighborhood of the previously predicted path parameter.

\subsubsection{Selecting the number of path segments}

As detailed in Section \ref{sec:preliminaries_global_reference_path}, the \emph{global} reference is composed of $M$ path segments. To lower the computational load, only $\eta\leq M$ path segments are used to generate the \emph{local} reference that is incorporated into the optimization problem. The number of path segments $\eta$ cannot be arbitrarily small, and there is a minimum number of segments to be selected to ensure the robot follows the reference path along a prediction horizon.
The number of path segments $\eta$ in the local reference path is a function of the prediction horizon length, the individual path segment lengths, and the speed of the robot at each time instance. We select a conservative $\eta$ by considering the maximum longitudinal velocity $v_{\max}$ and imposing that the covered distance is lower than a lower bound of the travelled distance along the reference path, namely,
\begin{equation}\label{eq:eta}
\underbrace{\tau \sum^{N-1}_{j = 0} v_{j}}_{\text{Traveled dist.}}
\!\leq\!\!
\tau N v_{\max}
\!\leq\!\!\!
\underbrace{\sum^{m + \eta}_{i = m + 1} ||\boldsymbol{p}^{\textrm{r}}_{i+1} - \boldsymbol{p}^{\textrm{r}}_i||}_{\text{Waypoints dist.}}
\!\leq \!\!\!\!\!
\underbrace{\sum^{m + \eta}_{i = m+1} s_i}_{\text{Ref. path length}},
\end{equation}
where $m$ is the index of the closest path segment to the robot, $\tau$ is the length of the discretization steps along the horizon, and $s_i$ the length of each path segment.



\subsubsection{Maintaining continuity over the local reference path}\label{sec:reference_path}

We concatenate the $\eta$ reference path segments into a differentiable local reference path $\boldsymbol{L}^r$, which will be tracked by the LMPCC, as follows 
\begin{equation}\label{eq:concatenation}
    \bar{\boldsymbol{p}}^r(\theta_k) = \sum^{m+\eta}_{i = m}
    \sigma_{i,+}(\theta_k) \sigma_{i,-}(\theta_k)\varsigma_i(\theta_k),
\end{equation}
where
$\sigma_{i,-}(\theta_k) = 1/(1+e^{(\theta - \sum^{i}_{j = m}s_i)/\epsilon})$
 and 
$\sigma_{i,+}(\theta_k) = 1/(1+e^{(-\theta + \sum^{i-1}_{j = m}s_i)/\epsilon})$
are two sigmoid activation functions for each path segment and $\epsilon$ is a small design constant. This representation ensures a continuous representation of the local reference path needed to compute the solver gradients.

\subsubsection{Cost function}\label{sec:cost_function}
\begin{figure}[t]
\centering
\includegraphics[trim={11.5cm 0 1.8cm 14.5cm},clip,width=.8\columnwidth]{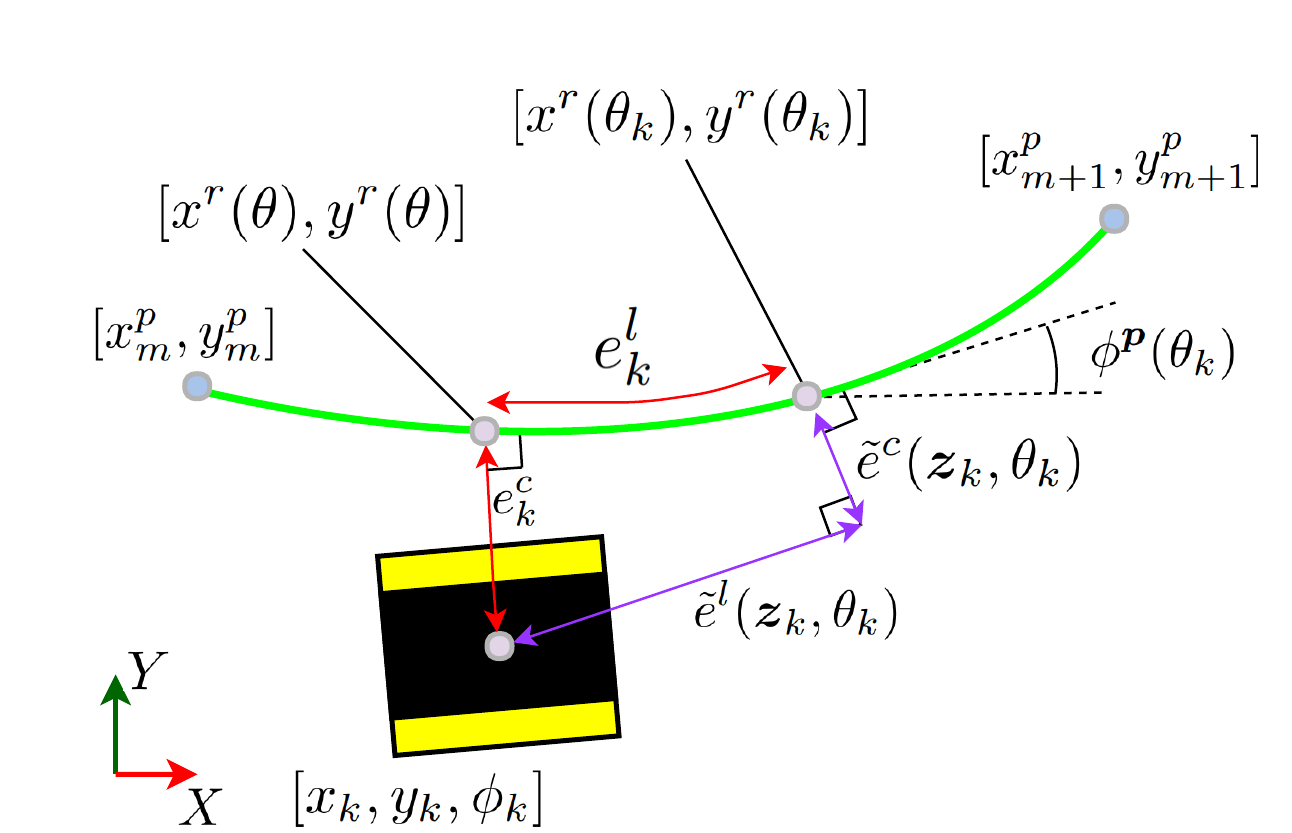}
\caption{Approximated contour and lag errors on the path segment.}
\label{fig:mpcc}
\end{figure}

 For tracking of the reference path, a contour and a lag error are defined, see \cref{fig:mpcc} and combined in an error vector $\boldsymbol{e}_k := [\tilde{\epsilon}^c(\boldsymbol{z}_k,\theta_k),\tilde{\epsilon}^l(\boldsymbol{z}_k,\theta_k)]^\textrm{T}$, with

\begin{align}
\boldsymbol{e}_k = 
\begin{bmatrix}
\sin \phi(\theta_k)& -\cos \phi(\theta_k)\\
- \cos \phi(\theta_k)&-\sin \phi(\theta_k)
\end{bmatrix}
\big( \boldsymbol{p}_k - \bar{\boldsymbol{p}}^r(\theta_k) \big),
\end{align}
where $\phi(\theta_k)=\arctan(\partial y^r(\theta_k)/\partial x^r(\theta_k)$ is {the direction of the path}.
Consequently, the LMPCC tracking cost is 
\begin{equation}\label{eq::tracking}
J_{\text{tracking}}(\boldsymbol{z}_k,\theta_k) = \boldsymbol{e}_k^TQ_{\epsilon}\boldsymbol{e}_k,
\end{equation}
where $Q_{\epsilon}$ is a design weight.

The solution which minimizes the quadratic tracking cost defined in Eq. \ref{eq::tracking} drives the robot towards the reference path. To make progress along the path we introduce a cost term that penalizes the deviation of the robot velocity $v_k$ from a reference velocity $v_{\textrm{ref}}$, i.e., $J_{\text{speed}}(\boldsymbol{z}_k,\boldsymbol{u}_k) = Q_v(v_{\text{ref}} - v_k)^2$ with $Q_v$ a design weight.
This reference velocity is a design parameter given by a higher-level planner and can vary across path reference segments.

To increase the clearance between the robot and moving obstacles, we introduce an additional cost term similar to a potential function,
\begin{equation}\label{eq:repulsive}
J_{\text{repulsive}}(\boldsymbol{z}_k) = Q_R \sum^{n}_{i = 1} \bigg( \frac{1}{(\Delta x_k)^2 + (\Delta y_k)^2 + \gamma} \bigg) \quad \!,
\end{equation}
 where $Q_R$ is a design weight, and $\Delta x_k$, $\Delta y_k$ represent the components of
 the distance from the robot to the dynamic obstacles. A small value $\gamma\geq 0$ is introduced for numerical stability. Eq. \ref{eq:repulsive} adds clearance with respect to obstacles and renders the method more robust to localization uncertainties.




Additionally, we penalize the inputs with $J_{\text{input}}(\boldsymbol{z}_k,\theta_k) = \boldsymbol{u}_k^TQ_{u}\boldsymbol{u}_k$, where $Q_u$ is a design weight.

The \ac{lmpcc} control problem is then given by a receding horizon nonconvex optimization, formally,
\begin{subequations}\label{eq:control_problem}
\begin{align}
  \!\!\!\!\!\! \!\!\! \!\!\!\!\!\!\!\!\! \!\!\! \!\!\!J^*\!\! =\!\! &\min_{\boldsymbol{z}_{0:N},\boldsymbol{u}_{0:N-1},\theta_{0:N}}\!\sum^{N-1}_{k =0}\!\! J(\mathbf z_k,\mathbf u_k, \theta_k)\!+\! J(\boldsymbol{z}_N,\theta_N) \!\\
\text{s.t.}:\, &~\textrm{\eqref{eq:dynamics_path_parameter}, \eqref{eq:state_control_constraints}},\\
&   \!\!\!c^{\text{stat},l,j}_k(\boldsymbol{z}_k) > 0, \, \forall j \in \{1,\dots,n_c \}, l \in \{1,...,4\} \\
&   \!\!\!c_k^{\textrm{obst},j}(\boldsymbol{z}_k) > 1, \, \forall j \in \{1,\dots,n_c \},\, \forall\, \textrm{obst}
\end{align}
\end{subequations}
where the stage cost is $ J(\mathbf z_k,\mathbf u_k, \theta_k): = J_{\text{tracking}}(\boldsymbol{z}_k,\theta_k) + 
J_{\text{speed}}(\boldsymbol{z}_k,\boldsymbol{u}_k) +
J_{\text{repulsive}}(\boldsymbol{z}_k) +
J_{\text{input}}(\boldsymbol{u}_k)$ 
and the terminal cost is $J(\boldsymbol{z}_N,\theta_N): = J_{\text{tracking}}(\boldsymbol{z}_N,\theta_N) + J_{\text{repulsive}}(\boldsymbol{z}_N)$. Eq. 11c and Eq. 11d are defined by Eq. \ref{eq:env_constraint2} and Eq. \ref{eq:ineq_constraint}, respectively.
Algorithm \ref{alg_lmpcc} summarizes our method. Note that at each control iteration, cThe selection of $\textrm{\texttt{iter}}_{\max}$ is empirically based on the maximum number of iterations allowed within the sampling time of our system to guarantee real-time performance.
\begin{algorithm}[t]
\caption{Local Model Predictive Contouring Control}\label{alg_lmpcc}
\begin{algorithmic}[1]
\State Given $\boldsymbol{z}_\textrm{init}$, $\boldsymbol{z}_{\textrm{goal}}$, $\pazocal{O}^{\textrm{static}}$, $\pazocal{O}_0^{\textrm{dyn}}$, and $N$
\For{$t = 0,1,2,...$}
    \State  $\boldsymbol{z}_0=\boldsymbol{z}_\textrm{init}$
    \State Estimate $\theta_0$ according to \cref{sec:path_parameter}
    \State Select $\eta$ according to \cref{eq:eta}
    \State Build $\bar{\boldsymbol{p}}^r(\theta_k)$, $k=1,...,N$, according to  \cref{eq:concatenation}
    \State $\text{Compute }c^{\text{stat,j}}_k(\boldsymbol{p}_k) \text{ along } \boldsymbol{q}_{0:N}$
    \State Get dynamic-obstacles predicted pose (Sec.~\ref{sec:dynamic_collision_avoidance})
      \State Solve the optimization problem of \cref{eq:control_problem}\label{alg:kkt}
    \State$\text{Apply } \boldsymbol{u^*_0}$
    \EndFor
\end{algorithmic}
\end{algorithm}


\section{RESULTS}\label{sec:results} 





This section presents experimental and simulation results for three scenarios with a mobile robot. We evaluate different settings for the parameters in our planner (\cref{sec:results_parameters}), as well as compare its performance in static (\cref{sec:results_static}) and dynamic (\cref{sec:results_dynamic}) environments against state of art motion planners \cite{Chang2017,fox1997dynamic,everett2018motion}.
A video demonstrating the results accompanies this paper.
\subsection{Experimental setup}\label{sec:experiment_setup}
\subsubsection{Hardware Setup}
\label{sec:results_hardware_setup}
Our experimental platform is a fully autonomous Clearpath Jackal ground robot, for which we implemented on-board all the modules used for localization, perception, motion planning, and control. Our platform is equipped with an Intel i5 CPU@2.6GHz, which is used to run the localization and motion planning modules, a Lidar Velodyne for perception, and an Intel i7 NUC mini PC to run the pedestrian tracker.
\subsubsection{Software Setup}
\label{sec:results_software_setup}

To build the global reference path we first define a series of waypoints and we construct a smooth global path by connecting the waypoints with a clothoid. We then sample intermediate waypoints from this global path and connect them with $3$rd order polynomials, which are then used to generate the local reference path.

The robot localizes with respect to a map of the environment, which is created before the experiments. For static collision avoidance the robot utilizes a map that is updated online with data from its sensors and we employ a set of rectangles to model the free space. Our search routine expands the sides of a vehicle-aligned rectangle simultaneously in the occupancy grid environment with steps of $\Delta^ {\text{search}}= 0.05 \text{m}$, until either an occupied cell is found or the maximum search distance $\Delta_{\text{max}}^{\text{search}} = 2 \text{m}$ is reached. Once an expanding rectangle side is fixed as a result of an occupied cell, the rest of the rectangle sides are still expanded to search for the largest possible area. This computation runs in parallel to the LMPCC solver, in a different thread, and with the latest available information.
Our experiments employ the open-source SPENCER Pedestrian tracker and 2d laser data \cite{linder2016multi} for detection and tracking of dynamic obstacles. If a pedestrian is detected, it is removed from the static map and treated as a moving ellipse. Our simulations use the open-source ROS implementation of the Social Forces model \cite{helbing1995social} for pedestrian simulation.

The LMPCC problem of Eq. \ref{eq:control_problem} is nonconvex. Our planner solves this problem online in real time using ACADO \cite{Houska2011a} and its \texttt{C}-code generation tool. We use a continuous-time kinematic unicycle motion model~\cite{Siegwart2011} to describe the robot's kinematics. The model is then discretized directly in ACADO using a multiple-shooting method combined with a Gauss-Legendre integrator of order 4, no Hessian approximations, and a sampling time of 50 ms. 
We select qpOASES \cite{ferreau2014qpoases} to solve the resulting QP problem and set a KKT tolerance of $10^{-4}$ and a maximum of 10 iterations.
If no feasible solution is found within the maximum number of iterations, then the robot decelerates. The planner computes a new solution in the next cycle. Based on our experience, this allowed to recover the feasibility of the planner quickly.
Our motion planner is implemented in \texttt{C++}/ROS and will be released open source. 




\begin{figure}[t!]
\centering
\begin{minipage}{0.4\columnwidth}
  \includegraphics[height=3.6cm,trim={0cm 0cm 0cm 0cm},clip]{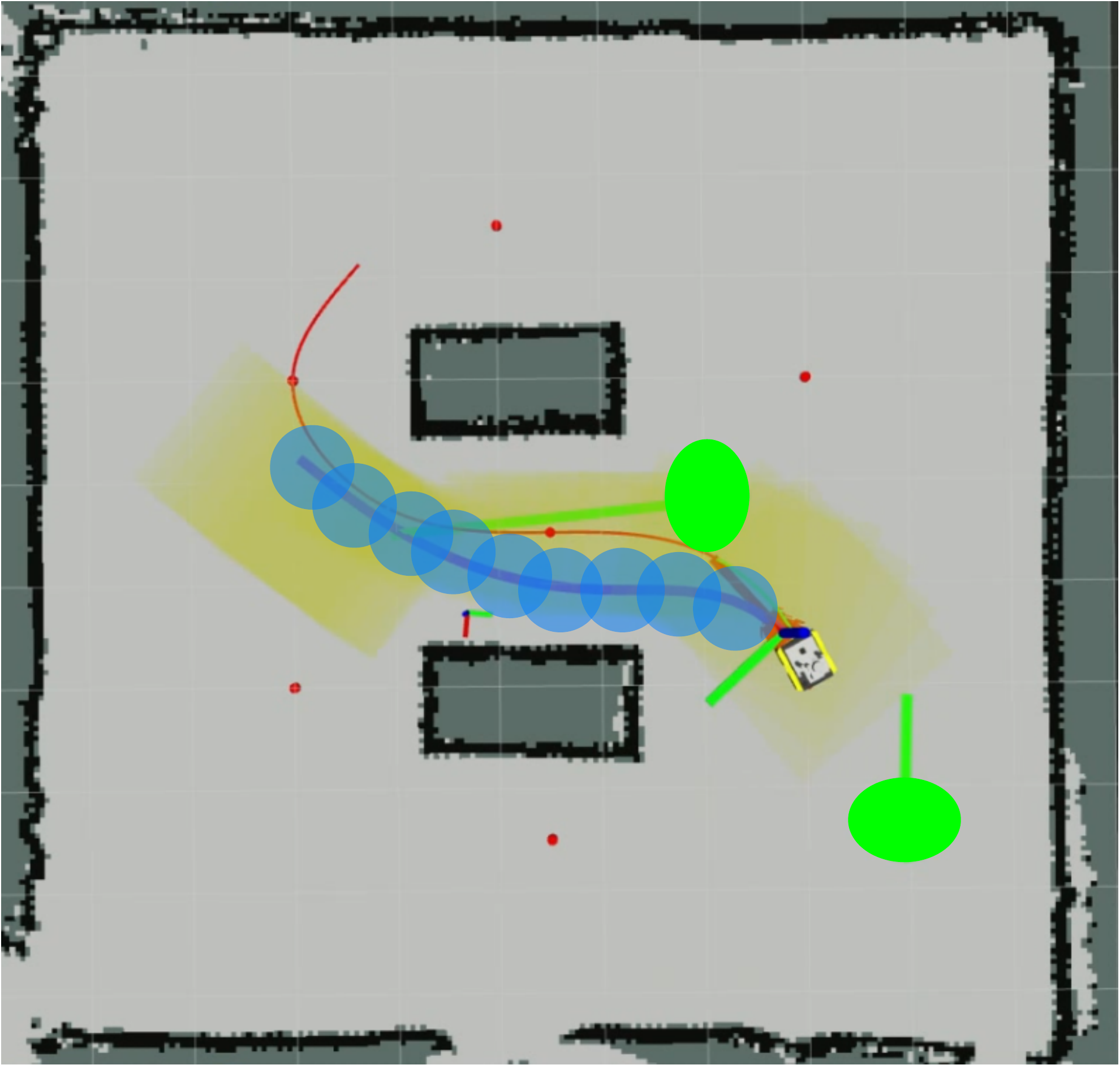}
  \caption*{\textit{a)} Visualization}
  \label{fig:test1} 
\end{minipage}
\hspace{1mm}
\begin{minipage}{0.4\columnwidth}
  \includegraphics[height=3.6cm,width=4.2cm,trim={1cm 1.4cm 0cm 1.3cm},clip]{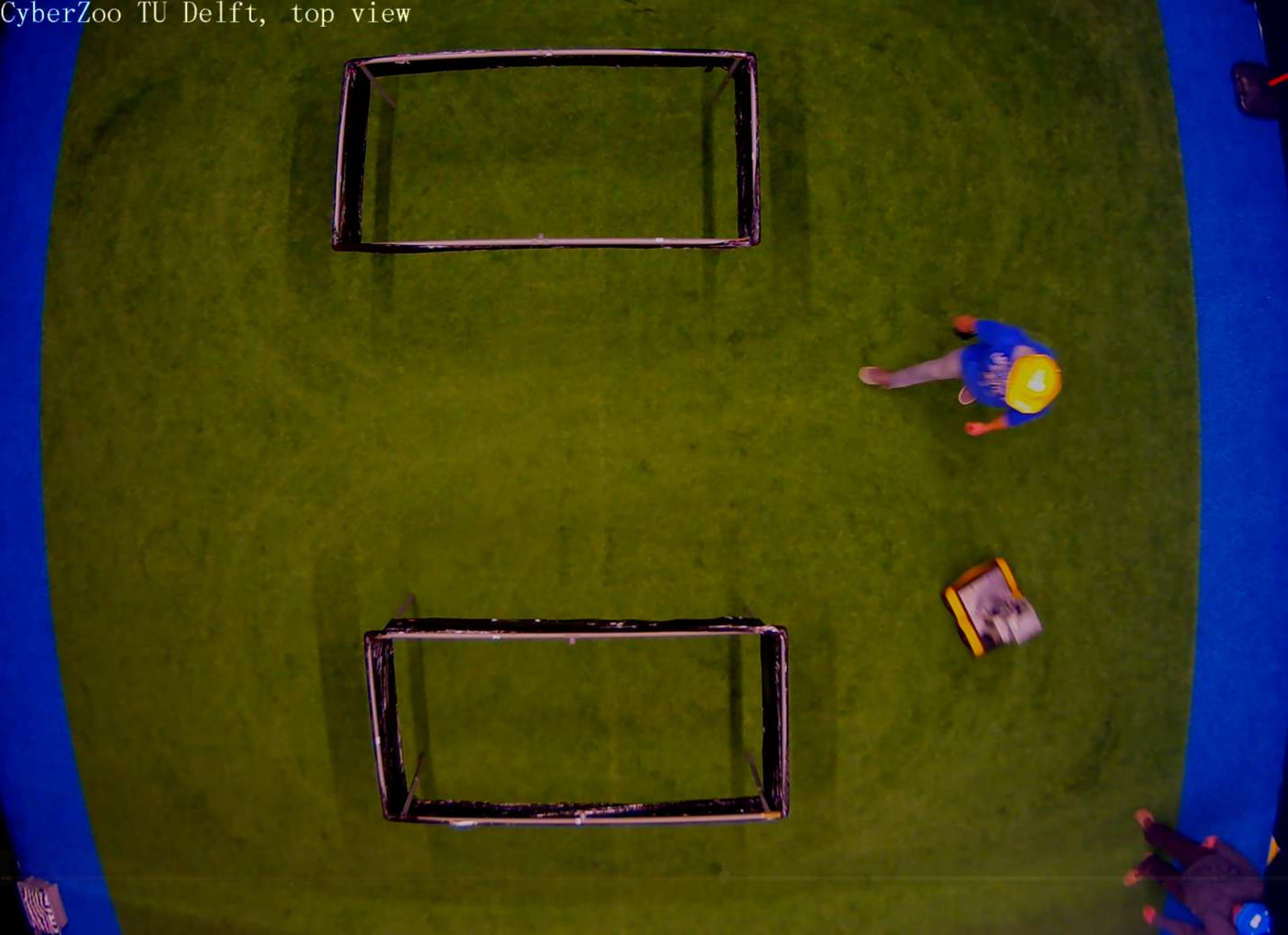}
  \caption*{\textit{b)} Top view}
  \label{fig:test2}
\end{minipage}
\caption{Experiment for evaluation of performance}
\label{fig:global_assessement_experiment}
\end{figure}   

\subsection{Parameter evaluation}\label{sec:results_parameters}

\begin{figure}[t]
\centering
\includegraphics[width=\columnwidth,trim={0cm 0cm 0cm 0cm}]{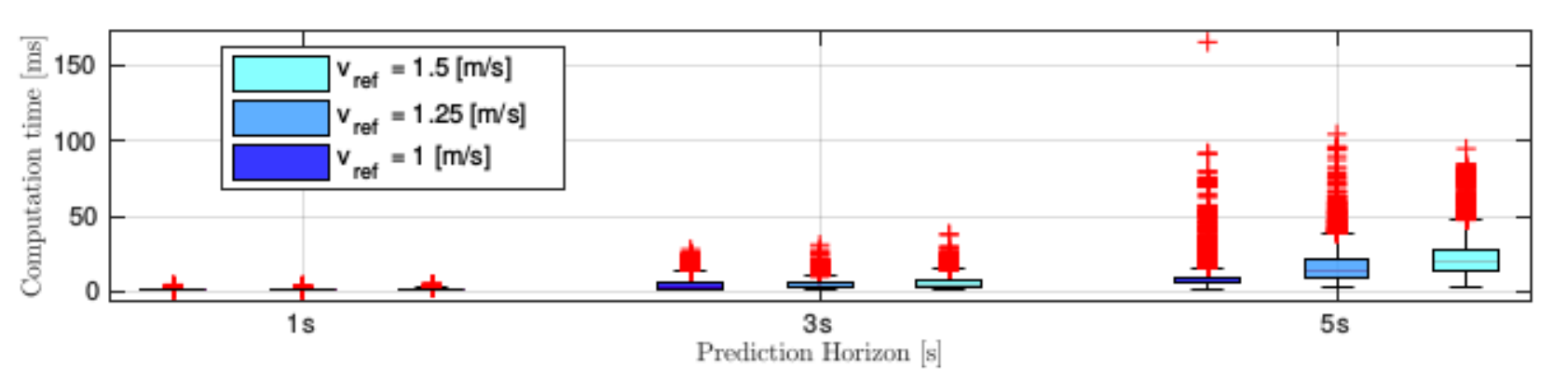}
\caption{Computation time required to solve the LMPCC problem online for different velocity references and horizon lengths. The central mark indicates the median. The bottom and top edges of the box indicate the 25th and 75th percentiles, respectively. The red crosses represent the outliers.}
\label{fig:computation_times}
\end{figure}

To evaluate the performance of the planner we performed several experiments at different reference speeds, $v_{\textrm{ref}}\in\{1\,\text{m/s}, 1.25\,\text{m/s}, 1.5\,\text{m/s}\}$, and prediction-horizon lengths, $T_{\textrm{Horizon}} \in \{1\,\text{s}, 3\,\text{s}, 5\,\text{s}\}$.
The robot follows a figure-8 path (red line in Figure~\ref{fig:global_assessement_experiment}-(a)) while avoiding two pedestrians (green ellipses) and staying within the collision-free area (yellow rectangles). We use one circle to represent the planned position of the robot (light blue circles). Each pedestrian is bounded by an ellipse of semi-axis 0.3 m and 0.2 m. The predicted positions of the pedestrians are represented by a green line. We align the semi-minor axis to the pedestrian walking direction.
For this experiment we rely on a motion capture system to obtain the position of the obstacles and robot. 
Figure~\ref{fig:computation_times} shows the computation time to solve the optimization problem.
For $v_{\textrm{ref}} \in\{ 1, 1.25, 1.5\}$ m/s and $T_{\textrm{Horizon}} \in\{ 1, 3\}$ s the computation times are under 50 ms, which is lower than the cycle-time defined for the planner. But, for $T_{\textrm{Horizon}} = 5$ s the 99th percentile is above the 50ms, not respecting the real-time constraint. The cases in which the planner exceeds the sampling time of the system are the situations in which the pedestrians suddenly step in front of the robot or change their direction of motion, requiring the solver more iterations to find a feasible solution. If not all the constraints can be satisfied, our problem becomes infeasible, and no solution is found. In this case, we reduce the robot velocity, allowing the solver to recover the feasibility after few iterations.
Table \ref{tab:clearance} summarizes the behavior of the planner in terms of clearance, that is, the distance from the border of the circle of the robot to the border of the ellipse defining the obstacles. It demonstrates that a short horizon leads to lower safety distances and more importantly, that our method was able to keep a safe clearance in most cases.

Based on these results, we selected a reference speed of 1.25 m/s and a horizon length of 3 s for the following experiments.


\begin{table}[]
\caption{Clearance between the robot and the dynamic obstacles for different velocity references and horizon lengths.}
\begin{tabular}{|c|c|c|c|}
\hline
\multirow{2}{*}{Prediction Horizon [s]} & \multicolumn{3}{c|}{Clearance Mean (1st percentile) [m]}  \\ \cline{2-4} 
                                        & $v_{ref}=1$ & $v_{ref}=1.25$ & $v_{ref}=1.5$ \\ \hline
1                                       &    1.54 (0.077)         &  1.60 (0.025)              &  1.69 (0.003)                \\ \hline
3                                       &    1.66 (0.077)         &  1.69 (0.072)              &  1.82 (0.065)                \\ \hline
5                                       &    1.69 (0.062)          & 1.68 (0.055)               & 1.92 (0.043)                  \\ \hline
\end{tabular}
\label{tab:clearance}
\end{table}
\subsection{Static collision avoidance}\label{sec:results_static}
In this experiment we compare the proposed planner with two baseline approaches:
\begin{itemize}
    \item \emph{A MPC tracking controller} \cite{Chang2017}. We minimize the deviation from positions on the reference path up to $1$ m ahead of the robot.
    \item \emph{The Dynamic Window (DW)} \cite{fox1997dynamic}. We use the open-source ROS stack implementation. The DW method receives the next waypoint once the distance to the current waypoint is less than 1 m.
\end{itemize}

For this and the following experiment, we fully rely on the onboard localization and perception modules. A VLP16 Velodyne Lidar is used to build and update a local map centered in the robot for static collision avoidance. The prebuilt offline map is only used for localization.

\begin{figure}
\centering
\includegraphics[width=0.7\columnwidth,trim={0cm 0cm 0cm 0cm}]{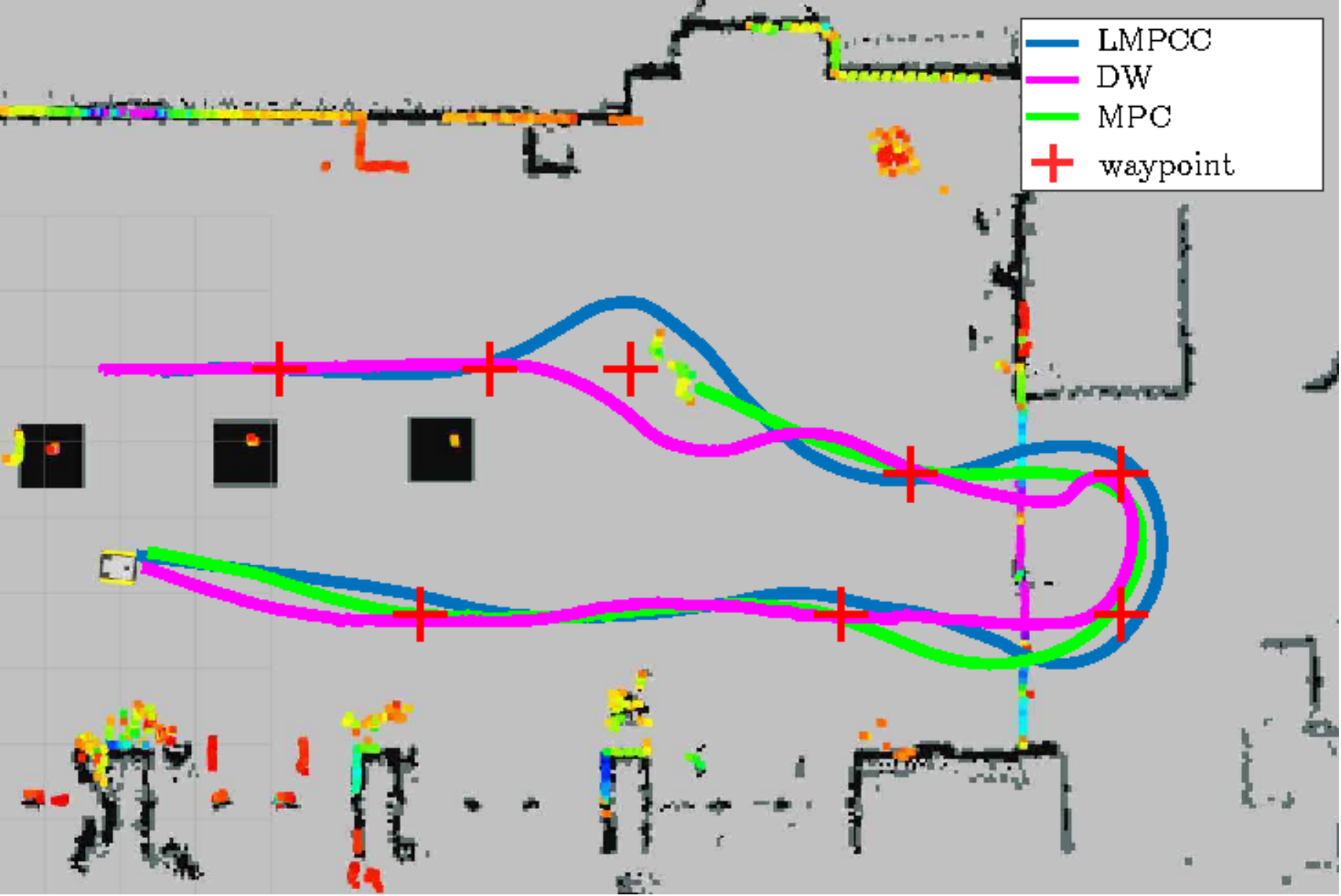}
\caption{Static collision avoidance scenario. The red crosses depict the path to follow (waypoints) and the colored points are the laser scan data. The blue, green, and magenta lines represent the trajectory obtained with the \ac{lmpcc},the MPC tracking controller, and the Dynamic Window, respectively.}
\label{fig:static_experiment}
\end{figure}

In this experiment the mobile robot navigates along a corridor while tracking a global reference path (red waypoints in Figure~\ref{fig:static_experiment}). When it encounters automatic doors, the robot must wait for them to open. When it encounters the obstacle located near the third upper waypoint from the left, which was not in the map, the robot must navigate around it. 
Figure~\ref{fig:static_experiment} shows the results of this experiment. All the three approaches are able to follow the waypoints and interact with the automatic door. When they encounter the second obstacle the classical MPC design fails to proceed towards the next waypoint. The DW and the LMPCC approaches are able to complete the task. Both methods showed similar performance (e.g., the traveled distance was 30.59 m for our LMPCC and 30.61 m for the DW). The time to the goal was relatively higher for the DW (50 s) compared to the LMPCC (41 s). This difference is mainly due to the ability of the LMPCC to follow a reference velocity. In addition, we tested DW and LMPCC in a scenario where half of the first door does not open due to malfunction. In such scenario, the LMPCC was able to traverse around the broken door while the DW gets into a deadlock state due to the narrow opening. A video of the experiments accompanies the paper.


\subsection{Dynamic collision avoidance}\label{sec:results_dynamic}

Our simulations compare our planner with an additional baseline: \emph{a socially-aware motion planner } named Collision Avoidance with Deep RL (CADRL) \cite{everett2018motion}. We employ the open-source ROS stack implementation. The CADRL method receives the same waypoints as the DW method.

\subsubsection{Simulation Results}

We compare the performance of the MPCC planner with the DW and CADRL baselines in the presence of pedestrians. Figure \ref{fig:front_page} shows the setup of our experiment. To avoid the overlap of the static and dynamic collision constraints, the detected pedestrians are removed from the updated map used for static avoidance and modeled as ellipses with a constant velocity estimate. The global path, as described in Section \ref{sec:experiment_setup},  consists of a straight line along the corridor. The robot has to follow this path while avoiding collisions with several pedestrians moving in the same or opposite direction. In this experiment, we do not evaluate the MPC tracking controller since it was unable to complete the previous experiment.
Aggregated results in \cref{tab:global_performance} show that the LMPCC outperforms the other methods. It achieves a considerably lower failure rate, smaller traveling distances, and maintains larger safety distances to the pedestrians. Only for the four pedestrians case, the DW achieved larger mean clearance, but with larger standard deviation. By accounting for the predictions of the pedestrians, our method can react faster and thus generate safer motion plans. \cref{tab:global_performance} also shows that the number of failures grows with the number of agents. Yet, our method can scale up to six pedestrians with low collision probability and perform real-time. For larger crowds our method would select the closer $6$ pedestrians.

\begin{table*}[]
 \caption{Statistic results of minimum distance to the pedestrians (where clearance is defined as the border to border distance), traveling distance and percentage of failures obtained for 100 random test cases of the dynamic collision avoidance experiment for $n \in \{2,4,6\}$ agents. The pedestrians follow the social forces model \cite{helbing1995social}.}
        \centering
\begin{tabular}{|p{1cm}|p{1.4cm}|p{1.4cm}|p{1.4cm}|p{1.4cm}|p{1.4cm}|p{1.4cm}|p{1.4cm}|p{1.4cm}|p{1.4cm}|p{1.4cm}|}
\hline
\multicolumn{2}{|c|}{\# agents} & \multicolumn{3}{c|}{Clearance Mean (1st percentile) [m]} & \multicolumn{3}{|c|}{\% failures (\% collisions / \% stuck) } & \multicolumn{3}{c|}{Traveled distance Mean (Std.) [m]} \\ \hline
\multicolumn{2}{|c|}{}        & DW                & CADRL               & LMPCC               & DW               & CADRL               & LMPCC              & DW              & CADRL             & LMPCC             \\ \hline
\multicolumn{2}{|c|}{2}                &   0.28 (0)                &      0.15 (0)            &   \textbf{0.29} (0.015)                  &  20 (19 / 1)               &4  (4 / 0)                   & \textbf{2} (0 / 2)               &    17.99 (3.9)           &     19.27 (3.7)              &     \textbf{15.81} (3.2)          \\ \hline
\multicolumn{2}{|c|}{4}                &       \textbf{0.46} (0)           &      0.32 (0)                &     0.25 (0.026)                &  35 (32 / 3)                &     31 (31 / 0)             &  \textbf{5} (2 / 3)                &     19.43 (5.6)            &      21.34 (4.4)             &      \textbf{15.77} (4.4)          \\ \hline
\multicolumn{2}{|c|}{6}                &      0.06 (0)             &       0.33 (0)               &    \textbf{0.38} (0.013)                 &     43 (41 / 2)             &     51 (51 / 0)             &  \textbf{7} (5 / 2)                &     21.09 (4.8)            &    18.97 (5.6)       &     \textbf{16.13} (1.9)            \\ \hline
\end{tabular}
 \label{tab:global_performance}
\end{table*}

\subsubsection{Experimental Results}
\begin{figure}[t]
\centering
\begin{subfigure}{\linewidth}
\includegraphics[width=\linewidth]{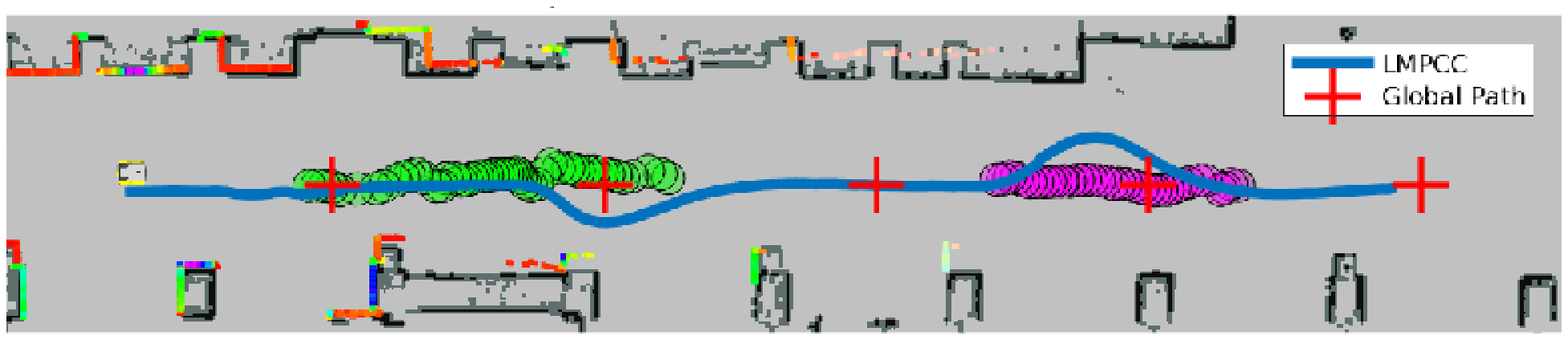}
\caption{LMPCC}
\end{subfigure}\quad
\begin{subfigure}{\linewidth}
\includegraphics[width=\linewidth]{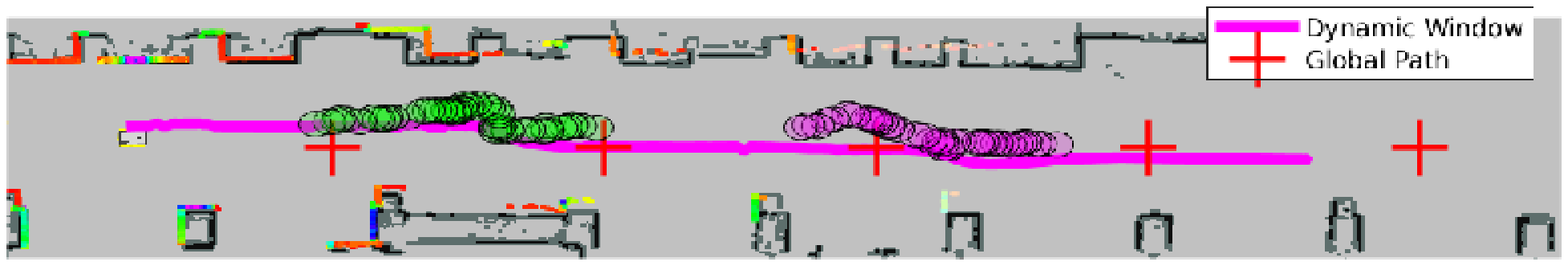}
\caption{Dynamic Window}
\end{subfigure}
\caption{Dynamic collision avoidance scenario. The red crosses represent the global path to follow (waypoints). The blue and magenta lines represent the trajectory executed by our \ac{lmpcc} (top) and by the Dynamic Window (bottom), respectively. The trajectories of the two pedestrians are represented by the green and magenta circles. In the lower case (dynamic window) the robot reacts late and the pedestrians must actively avoid it.}\label{fig:dynamic_experiment}
\end{figure}

We use the previous setup to compare the LMPCC and DW methods on a real scenario with two pedestrians. We do not test the CADRL method because the current open-source implementation does not allow static collision avoidance for unconstrained scenarios such as the faculty corridor depicted in Figure \ref{fig:dynamic_experiment}. Figure \ref{fig:dynamic_experiment} shows one representative run of our method (top) and the DW (bottom). We observe that the proposed method reacts in advance to avoid the pedestrians, resulting in a larger clearance distance. In contrast, the DW reacts late to avoid the pedestrian, which has to avoid the robot himself actively. Our proposed method was able to navigate safely in all of our experiments with static and two pedestrians.

\subsection{Applicability to an autonomous car}
\label{sec:results_simulations}
To validate the applicability of our method to more complex robot models, we have performed a simulation experiment with a kinematic bicycle model \cite{kong2015kinematic} of an autonomous car.
The planner commands the acceleration and front steering.
The car follows a global reference path while staying within the road boundaries (i.e., the obstacle-free region) and avoiding moving obstacles (such as a simulated cyclist proceeding in the direction of the car and a pedestrian crossing the road in front of the car). The accompanying video shows the results where the autonomous vehicle successfully avoids the moving obstacles, while staying within the road limits. We refer the reader to \cite{Ferranti2019} for more details and results.
\section{CONCLUSIONS \& FUTURE WORK}\label{sec:conclusions}
This paper proposed a local planning approach based on Model Predictive Contouring Control (MPCC) to safely navigate a mobile robot in dynamic, unstructured environments.
Our local MPCC relies on an upper bound of the Minkowski sum of a circle and an ellipse to safely avoid dynamic obstacles and a set of convex regions in free space to avoid static obstacles.
We compared our design with three baseline approaches (classical MPC, Dynamic Window, and CADRL). The experimental results demonstrate that our method outperforms the baselines in static and dynamic environments. Moreover, the light implementation of our design shows the scalability of our method up to six agents and allowed us to run all algorithms on-board. Finally, we showed the applicability of our design to more complex robots by testing the design in simulation using the model of an autonomous car.
As future work, we intend to expand our approach for crowded scenarios, by accounting for the interaction effects between the robot and the other agents.

\bibliographystyle{IEEEtran}
\bibliography{MyBib}

\end{document}